\pdfoutput=1
\documentclass[11pt,a4paper]{article}
\usepackage[dvipsnames]{xcolor} 

\usepackage{authblk}
\usepackage[hyperref]{emnlp2021}

\usepackage{times}
\usepackage{latexsym}

\usepackage{microtype}

\usepackage{microtype}
\usepackage{booktabs}
\usepackage{url}
\usepackage{multirow}
\usepackage{adjustbox}

\usepackage{graphicx}
\graphicspath{ {images/} }
\usepackage{amsmath,amsfonts,amsthm,bm}
\usepackage{tabularx}
\usepackage{xspace}
\usepackage{soul}
\usepackage{xcolor}

\usepackage[colorinlistoftodos]{todonotes}

\newcommand{\namens}{JoGANIC}
\newcommand{\name}{\namens\xspace}
\newcommand{\fullnamens}{Journalistic Guidelines Aware News Image Captioning}
\newcommand{\fullname}{\fullnamens\xspace}

\newcommand{\tellns}{Tell}

\newcommand{\tellfullns}{Transform and Tell}
\newcommand{\tellfull}{\tellfullns\xspace}

\newcommand{\neeshort}{NEE\xspace}
\newcommand{\neeshortns}{NEE}

\newcommand{\good}{GoodNews\xspace}
\newcommand{\nyt}{NYTimes800k\xspace}
\title{\fullnamens}

\author[1]{Xuewen Yang}
\author[2]{Svebor Karaman}
\author[2]{Joel Tetreault}
\author[2]{Alex Jaimes}
\affil[1]{Stony Brook University}
\affil[2]{Dataminr Inc.}
\affil[1]{\tt xuewen.yang@stonybrook.edu}
\affil[2]{\tt \{skaraman, jtetreault, ajaimes\}@dataminr.com}

\date{}

\begin{document}
\maketitle
\begin{abstract}
The task of news article image captioning aims to generate descriptive and informative captions for news article \textit{images}. 
Unlike conventional image captions that simply describe the content of the image in general terms, news image captions follow journalistic guidelines and rely heavily on named entities to describe the image content, often drawing context from the whole article they are associated with. 
In this work, we propose a new approach to this task, motivated by caption guidelines that journalists follow.
Our approach, \fullname (\name), leverages the 
structure of captions to improve the generation quality and guide our representation design.
Experimental results, including detailed ablation studies, on two large-scale publicly available datasets show that \name substantially outperforms state-of-the-art methods both on
caption generation and named entity related metrics.
\end{abstract}



\section{Introduction}
Research on generating textual descriptions of images has made great progress in recent years with the introduction of encoder-decoder architectures~\cite{Kelvin2015,Johnson16,Venugopalan2017,Karpathy2017,Anderson2018,Lu2018NeuralBT,Aneja18}.
Those models are generally trained and evaluated on image captioning datasets like COCO~\cite{Lin14,ChenCOCO15} and Flickr~\cite{Hodosh13} that only contain generic object categories but no details such as names, locations, or dates.    
The captions generated by these methods are thus generic descriptions of the images.

\begin{figure}[t]
    \centering
    \includegraphics[width=\columnwidth]{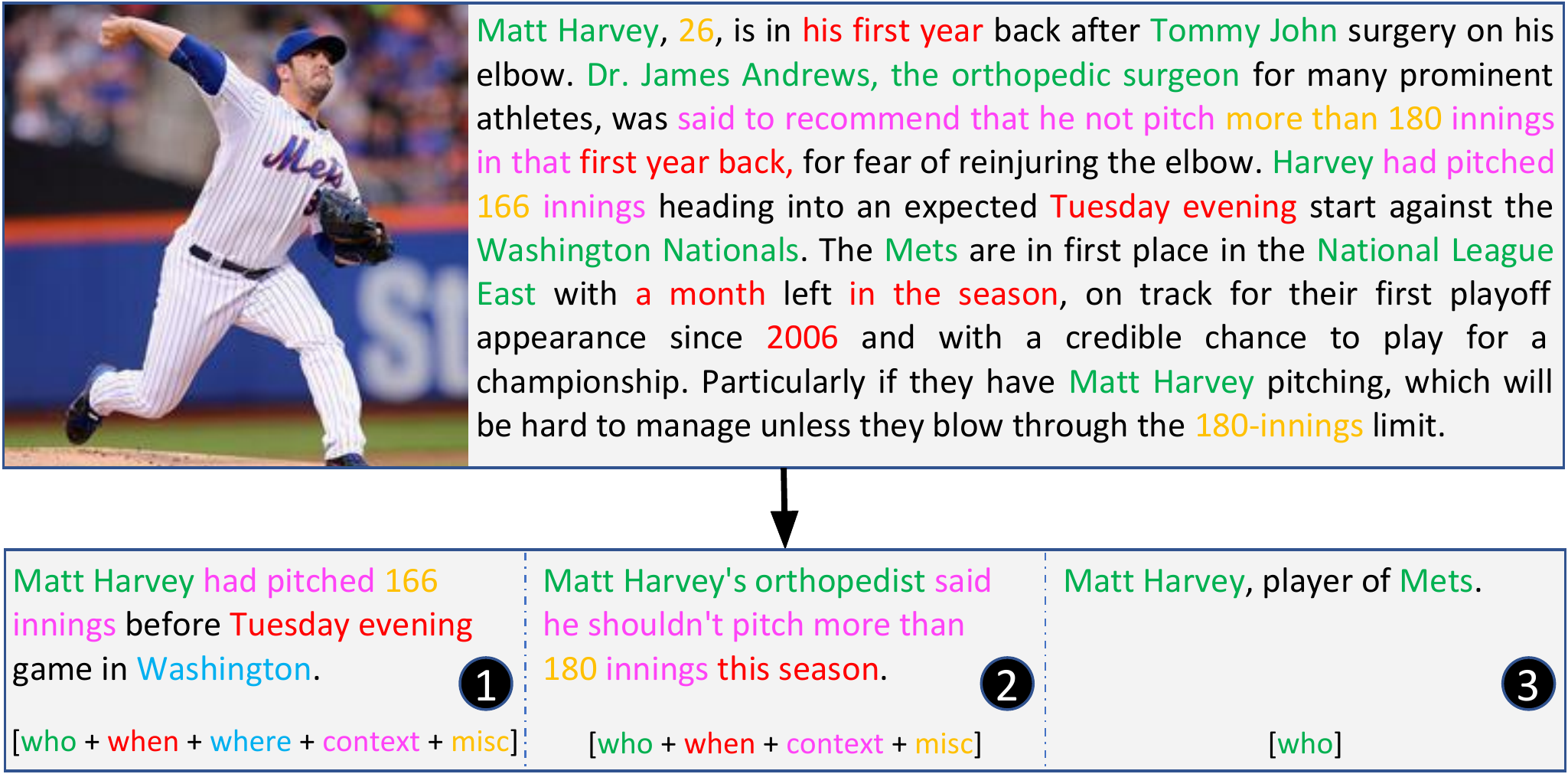}
    \caption{Three possible captions (bottom) for one image-article pair input (top).
    These three captions follow different `templates' composed of 
    \textit{who} ({in \color{ForestGreen}green}),
    \textit{when} ({in \color{red}red}), 
    \textit{where} ({in \color{Cerulean}blue}),
    \textit{context} ({in \color{VioletRed}purple}) and \textit{misc} ({in \color{YellowOrange} orange}) components.}
    \label{fig:captioning}
    \vspace{-0.5cm}
\end{figure}

The news image captioning
problem~\cite{Feng13,Ramisa18,Biten_2019_CVPR,Tran2020Tell} can be seen as a multi-modal extension of the image captioning task with
additional context provided in the form of a news article.
Specifically, given image-article pairs as input, the news captioning task aims to generate an informative caption that describes the image with proper named entities and context extracted from the article.
The development of automatic news image caption generation methods can ease the process of adding images to articles and produce more engaging content.
According to \textit{The News Manual}\footnote{\url{https://www.thenewsmanual.net/Manuals\%20Volume\%202/volume2_47.htm}} and \textit{International Journalists' Network}\footnote{\url{https://ijnet.org/en/resource/writing-photo-captions}}, a caption should
help news readers understand six main components (\textit{who}, \textit{when}, \textit{where}, \textit{what}, \textit{why}, \textit{how}) related to the image and 
article.
As shown in Fig.~\ref{fig:captioning}, different journalists can write captions to cover different components for the same image and article pair.
Previous news image captioning work~\cite{Biten_2019_CVPR,Tran2020Tell} has not directly addressed the challenge of generating a  caption that follows those journalistic principles.


In this work, we tackle the news image captioning problem by introducing these guidelines in our modeling through a new
concept called a `caption template', which is composed of $5$ key components, detailed in Section~\ref{sec:template}.
We propose a \fullname (\name) model that, given an image-article pair, aims to predict the most likely active template components and, using component-specific decoding block, 
 produces a caption following the provided template guidance.
 \name thus  
 models the underlying structure of the captions, which helps to improve the generation quality. 



Captions for images that accompany news articles often include named entities \textit{and} rely heavily on context found throughout the article (making the text encoding process especially challenging). We propose two techniques to address these issues: (i) integration of features specifically to extract relevant named entities, 
and (ii) a multi-span text reading (MSTR) method, which first splits long articles into multiple text spans and then merges the extracted features of all spans together.

Our work has two main contributions: (i) the definition of the template components of a news caption based on journalistic guidelines, and their explicit integration in the caption generation process of our 
\name model; 
(ii) the design of encoding mechanisms to extract relevant information for the news image captioning task throughout the article, specifically a dedicated named entity representation and the ability to process longer article. Experimental results show better performance than state of the art on news image caption generation. We release the source code of our method at \url{https://github.com/dataminr-ai/JoGANIC}.

\section{Related Work}



\subsection{Generic Image Captioning}
State-of-the-art approaches~\cite{Johnson16,wang2020unique,He2020image,Sammani_2020_CVPR}
mainly use encoder-decoder frameworks with attention to generate captions for images.
\citet{Kelvin2015} developed soft and hard attention mechanisms to focus on different regions in the image when generating different words.
Similarly, \citet{Anderson2018} used a Faster R-CNN~\cite{Ren15} to extract regions of interest that can be attended to.
\citet{Yang2020Fashion} used self-critical sequence training for image captioning.
\citet{Lu2018entity} and \citet{whitehead2018} introduced a knowledge aware captioning method where the knowledge comes from metadata associated with the datasets.

Our work differs from generic image captioning in 
three 
aspects: (i) our model's input consists of image-article pairs; 
(ii) our caption generation is a guided process following news image captioning journalistic guidelines;
(iii) news captions contain named entities and additional context extracted from the article, making them more complex. 


\subsection{News Article Image Captioning}

\begin{table}[t]
    \centering
    \small
    \begin{adjustbox}{max width=\columnwidth}
    \begin{tabular}{ccc}
    \toprule
    Type & Description & Component \\
    \midrule
    PERSON & People, including fictional & who\\
    NORP & Political groups & who \\
    ORG & Companies, agencies, etc &  who \\
    DATE & Dates or periods & when \\
    TIME & Times smaller than a day & when \\
    FAC & Buildings, airports, highways & where \\
    GPE & Countries, cities, states  & where \\
    LOC & Locations, mountains, waters & where  \\
    PRODUCT & Objects, vehicles, foods & misc \\
    EVENT & Named wars, sports events & misc \\
    ART & Titles of books, songs & misc \\
    LAW & Laws & misc \\
    LAN & Any named language & misc \\
    PERCENT & Percentage, including ``\%'' & misc \\ 
    MONEY & Monetary values & misc \\
    QUANTITY & Measurements & misc \\
    ORDINAL & ``first'', ``second'', etc & misc \\
    CARDINAL & Numerals & misc \\
    \bottomrule
    \end{tabular}
    \end{adjustbox}
    \caption{Named Entities type, description and assigned component category.}
    \label{tab:ne_p}
    \vspace{-0.3cm}
\end{table}


One of the earliest works in news article image captioning,~\citet{Ramisa18}, 
proposed an encoder-decoder architecture with a deep convolutional model VGG~\cite{Simonyan15} and  Word2Vec~\cite{Mikolov13} as the image and text feature encoder, and an LSTM as the decoder.

\citet{Biten_2019_CVPR} 
 introduced the GoodNews dataset,
 and
proposed a two-step caption generation process using ResNet-152~\cite{Kaiming16} as the image representation and a sentence-level aggregated representation using GloVe embeddings~\cite{pennington2014glove}.
First, a caption is generated with 
placeholders for the different types of named entities: \textit{PERSON}, \textit{ORGANIZATION}, \textit{etc.} shown in the left column of Table \ref{tab:ne_p}.
Then, the placeholders are filled in by matching entities from the best ranked sentences of the article.
This two-step process aims to deal with rare named entities but prevents the captions from being linguistically rich and is can induce error propagation between steps.  


More recently, \citet{Liu2020VisualNewsB, Hu2020, Tran2020Tell} proposed one step, end-to-end methods. 
They all used ResNet-152 as image encoder, while for the text encoder: \citet{Hu2020} applied BiLSTM, \citet{Liu2020VisualNewsB} used BERT and \citet{Tran2020Tell} used RoBERTa.
\citet{Hu2020, Liu2020VisualNewsB} used LSTM as the decoder. 
\citet{Tran2020Tell} introduced the \nyt dataset, 
and a model named \tellfull, which we refer to as \tellns. This model exploits a Transformer decoder and byte-pair-encoding (BPE)~\cite{sennrich2016neural} allowing to generate captions with unseen or rare named entities from common tokens.
As in other multimodal tasks, where studies~\cite{shekhar2019,caglayan2019,li2020multimodal} have shown that the exploitation of both modalities is essential for achieving a good performance, \citet{Tran2020Tell} evaluated a text only model showing that it performs worse than the multimodal model.
We will also evaluate single visual and text modality models in our experiments.
Our work differs from previous work in news image captioning in that \name is an end-to-end framework that (i) integrates journalistic guidelines through a template guided caption generation process; and (ii) exploits a dedicated named entity representation and a long text encoding mechanism.
Our experiments show that our framework significantly outpeforms the state of the art.


\section{Defining Caption Templates}
\label{sec:template}

\begin{table}[t]
    \centering
    \small
    \begin{adjustbox}{max width=\columnwidth}
    \begin{tabular}{cccccc}
    \toprule
      &  who  & when & where & misc & context \\
       \hline
      \good  &  93.02 & 44.06  & 58.59  & 31.69 &  78.44 \\
      \nyt  & 93.77 & 41.54 & 51.08  & 30.92 & 77.07 \\
       \bottomrule
    \end{tabular}
    \end{adjustbox}
    \caption{Template components percentage in the \good and \nyt datasets.}
    \label{tab:tc}
    \vspace{-0.3cm}
\end{table}

The objective of news image captioning is to give the reader a clear understanding of the main components \textit{who}, \textit{when}, \textit{where}, \textit{what}, \textit{why} and \textit{how} depicted in the image given the context of the article. 
We propose to exploit the idea of components in the caption generation process, but we first need to define components that can be automatically detected in the ground truth caption for training. 

The \textit{who}, \textit{when} and \textit{where} components can be retrieved via Named Entity Recognition (NER).
As shown in the right column of Tab.~\ref{tab:ne_p}, we define named entities with type `PERSON', `NORP' and `ORG' as \textit{who}, those with type `DATE' and `TIME' as \textit{when}, and ones with type `FAC', `GPE' and `LOC' as \textit{where}. 
We define the component \textit{misc} as the rest of the named entities.
The \textit{what}, \textit{why} and \textit{how} components are hard to define and can correspond to a wide range of elements, we propose to
merge them into a \textit{context} component,
which we assume 
is present if a verb is detected by a part-of-speech (POS) tagger\footnote{We use spaCy which has almost SOTA pos tagging accuracy of $97.8\%$ and NER accuracy of $89.8\%$ on the OntoNotes corpus.}
. 
In Fig.~\ref{fig:captioning}, captions 1 and 2 have an \textit{context} component, but caption 3 does not contain a verb and thus has no \textit{context}.

In summary, our proposed news caption template consists of \textit{at most} five components: \textit{who}, \textit{when}, \textit{where}, \textit{context} and \textit{misc}.
We report the percentage of each component in the captions of the \good and \nyt datasets in Tab.~\ref{tab:tc}. The \textit{who} is present in almost all the captions, and all components appear 
commonly in both datasets.

\section{Template-Guided News Image Captioning}

In this section, we formally define the news captioning task and introduce the idea of template guidance and our \fullname (\name) approach.
We then propose two strategies 
to address the specific challenges 
of
named entities and long articles. 

\subsection{News Captioning Problem Formulation}
Given an image and article pair ($X^I$, $X^A$), the objective of news captioning is to generate a sentence $\mathbf{y}=\{\mathbf{y}_1,\ldots ,\mathbf{y}_N \}$ with a sequence of $N$ tokens, $\mathbf{y}_i \in V^K$ being the $i$-th token, $V^K$ being the vocabulary of $K$ tokens.
The problem can be solved by an encoder-decoder model. 
The decoder predicts the target sequence $\mathbf{y}$ conditioned on the source inputs $X^I$ and $X^A$.
The decoding probability $P(\mathbf{y}\vert X^I, X^A)$ is modeled using the probability of each target token $\mathbf{y}_n$ at time step $n$ conditioned on the source input $X^I$ and $X^A$ and the current partial target sequence $\mathbf{y}_{<n}$:
\begin{equation}
\resizebox{0.89\hsize}{!}{%
$P(\mathbf{y}\vert X^I, X^A;\bm{\theta}) = \prod_{n=1}^{N} P(\mathbf{y}_n\vert X^I, X^A, \mathbf{y}_{<n};\bm{\theta})$%
}
\end{equation}
where, $\bm{\theta}$ denotes the parameters of the model.

\subsection{Template Guidance}
\label{sec:template-guidance}
To make our model capable of generating captions following different templates, we introduce a new variable $\bm{\alpha}$
for template guidance.
The new decoding probability can be defined as:
\begin{equation}
\resizebox{0.89\hsize}{!}{
$P(\mathbf{y}\vert X^I, X^A) = \prod_{n=1}^{N} P(\mathbf{y}_n\vert X^I, X^A, \bm{\alpha}, \mathbf{y}_{<n})$
}
\end{equation}
where we ignore $\bm{\theta}$ for simplicity.

Based on our definition of templates, 
we could see $\bm{\alpha}$ as the high-level template class
defined by the combination of the active components. 
As there are $5$ template components, the total number of possible template classes is $2^5$.
However, this  poses two challenges to train our model: 
(i) data imbalance, as the most frequent template corresponds to 15.2\% of captions, while the least common ones appear less than 2\% of the time (more details in Tab. 3 of the supplementary material), and 
(ii) different high-level templates may be similar (i.e. having a single component difference) but would be considered totally different classes.

In order to address these issues we define $\bm{\alpha}$ as the set of active components of the template $\bm{\alpha}_{i=1}^5$, with $\bm{\alpha}_i$ being the probability of a template having component $i$. This formulation enables us to exploit the partial overlap in terms of components between the different templates.
Note that the percentage of each component, in Tab.~\ref{tab:tc}, is not as imbalanced as the full template classes. 
The template guidance $\bm{\alpha}$ can be provided by the news writer (`oracle' setting in the experiments) or can be estimated 
(`auto' setting) 
through a multi-label classification task as detailed in the next section and illustrated in the top-left of Fig.~\ref{fig:arch}(a). 

The template guidance variable $\alpha$ is static during the decoding process but that does not prevent our method from generating fluent captions covering the whole set of components. Exploring ways of exploiting dynamically the component specific representations during the caption generation process could be an interesting future work direction. 


\subsection{Our Model Description}
\begin{figure*}
    \centering
    \includegraphics[width=0.95\textwidth]{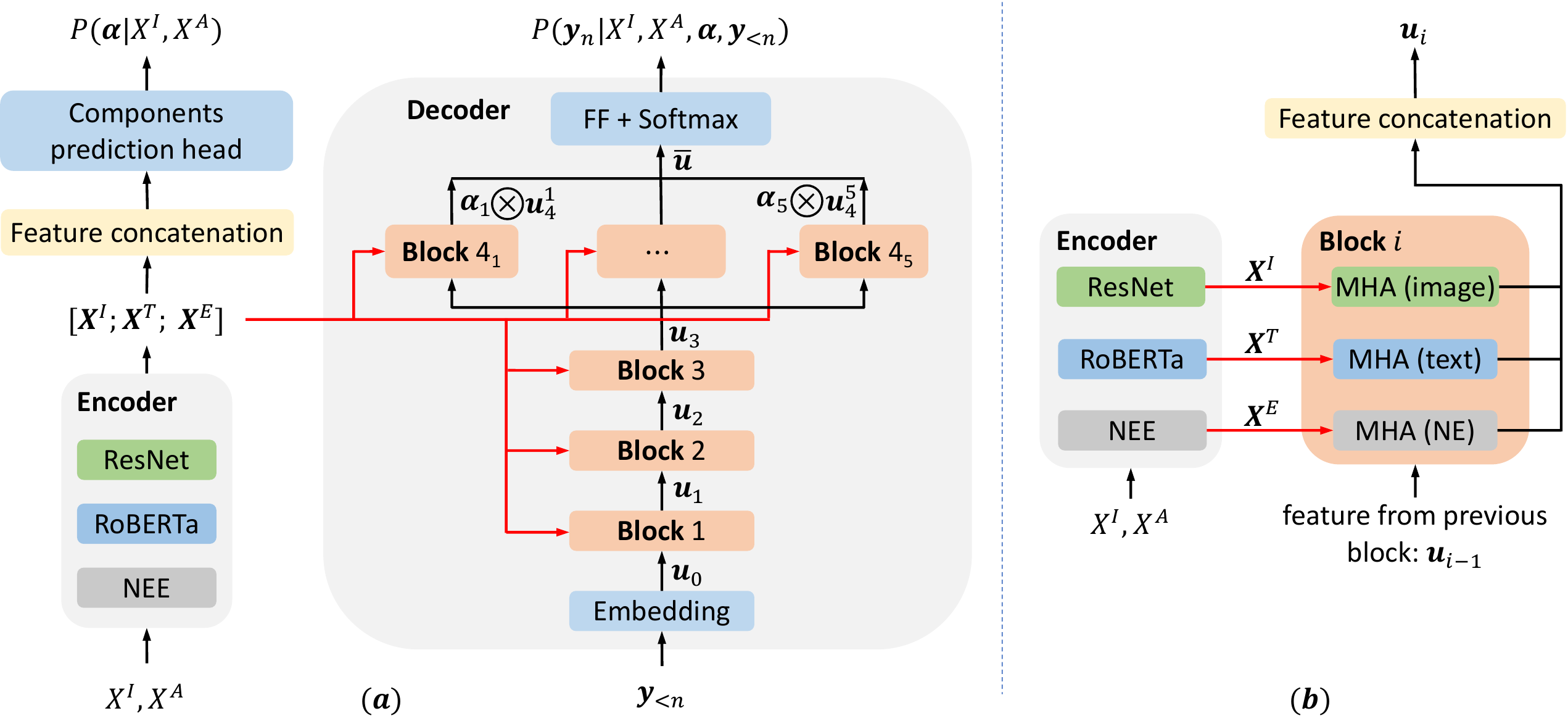}
    \caption{The architecture of our model. (\textit{a}) The Encoder takes image+text+named entities as input and generates features. The Decoder consists blocks 1-4, with blocks 1-3 shared for all template components \textit{who}, \textit{when}, \textit{where}, \textit{context} and \textit{misc}. Block 4 consists of 5 component-specific subblocks ($4_1$-$4_5$). A prediction head on top of the encoder predicts the probabilities of the $5$ components $\bm{\alpha}_{1:5}$, which then multiply the representations of the 5 subblocks $\mathbf{u}^{1:5}_4$. The final representation $\overline{\mathbf{u}}$ is obtained by averaging and used to predict the output token probabilities. (\textit{b}) Every block takes as input the representations from previous blocks as well as those from the Encoder via three Multi-Head Attention (MHA) modules designed for image, text and named entities separately.  
    }
    \label{fig:arch}
    \vspace{-0.33cm}
\end{figure*}

We propose a news image captioning model that generates captions through template guidance and can also generate accurate named entities and cover a larger extent of the article.
Our \name model, illustrated in Fig.~\ref{fig:arch}, is a transformer-based encoder-decoder, 
with an encoder extracting features from the image $X^I$ and the article $X^A$,
a prediction head estimating the probability of each component 
and a hybrid decoder to produce the caption.

The encoder consists of three parts: 
(i) a ResNet-152 pretrained on ImageNet extracting the image feature $\mathbf{X}^I \in \mathbb{R}^{d_I}$; 
(ii) RoBERTa producing the text features $\mathbf{X}^T \in \mathbb{R}^{d_T}$ from the article; and 
(iii) a Named Entity Embedder (NEE), detailed in Section~\ref{sec:nee}, applied to obtain the features  $\mathbf{X}^E \in \mathbb{R}^{d_E}$ of the named entities in the article.
The components prediction head, taking as input the concatenation of the image, article and named entities features, is a multi-layer perceptron with a sigmoid layer trained (using the components detected in the ground truth caption as target) to output the probability of each component $P(\bm{\alpha}\vert X^I,X^A)$.

The hybrid decoder consists of an embedding layer to get the embeddings of the output generated thus far (i.e., the partial generation), followed by $4$ blocks of $3$ Multi-Head Attention (MHA) modules, denoted as MHA (image/text/NE), to compute the attention across the partial generation and the input image, text and named entities. The final representation $\mathbf{u}_i$ for each block is the concatenation of the $3$ modules' output, Fig.~\ref{fig:arch}(b).
The first $3$ blocks are shared for all components, while
the $4$-th block consists of $5$ parallel component-specific blocks $4_1-4_5$ where block $4i$ outputs the representation $\mathbf{u}_4^i$ 
for the component $i$.
The final representation of the decoder is the average of the weighted sum of all components $\overline{\mathbf{u}}=\frac{1}{5}\sum_{i=1}^5\bm{\alpha}_i\mathbf{u}^i_4$.
Then the output probability $P(\mathbf{y}_{n}\vert X^I,X^A,\bm{\alpha},\mathbf{y}_{<n})$ is obtained by applying a feed-forward (FF) layer, and softmax over the target vocabulary.
Note that our ``template guided'' generation does not limit the number of occurrences of one component in the output caption 
and does not explicitly constrain the generation of specific components but rather the final representation $\overline{\mathbf{u}}$ will rely more on the component-specific representations corresponding to higher $\bm{\alpha}_i$ values. 


\subsubsection{Named Entity Embedding}
\label{sec:nee}
With over $96\%$ 
(see Tab. 1 in the supplementary material) of the news captions containing named entities, producing accurate named entities is essential to generating  good news captions.
However, text encoders like RoBERTa cannot properly represent named entities, and only handle them implicitly through BPE (Byte-Pair Encoding) subwords.

To deal explicitly with named entities, we learn entity embeddings from the Wikipedia knowledge base (KB),
following Wikipedia2vec~\cite{Wikipedia2Vec18} 
which embeds words and entities into a common space\footnote{\url{https://wikipedia2vec.github.io/wikipedia2vec/}}.
Given a vocabulary of words $V_{W}$ and a set of entities $V_{E}$, it learns a lookup embedding function $\mathcal{E}_{Wiki}:V_{W}\cup V_{E}\rightarrow \mathbb{R}^{d_{Wiki}}$.
There are three components in Wikipedia2Vec: (i) a skip-gram model for learning the word similarity in $V_{W}$, (ii) a KB graph model to learn the relatedness between pairs of entities  (vertices $V_{E}$ of the Wikipedia entity graph) and (iii) a version of Word2Vec where words are predicted from entities. 

Since predicting the correct named entities from context is very important for news captioning, we introduce a fourth component: (iv) a neural entity predictor (NEP).
Given a text (sequence of words) $t=\{w_1,\ldots,w_N \}$, we train Wikipedia2vec to predict the entities ${e_1,\ldots,e_m}$ that appear in the sequence.
With $E_{KB}$ being the set of all entities in KB, and $v_e$ and $v_t$ (computed as the element-wise mean of all the word vectors in $t$ followed by a fully connected layer) the vector representations of the entity $e$ and the text $t$, respectively, the probability of an entity $e$ appearing in text $t$ is defined as
\begin{equation}
    P(e\vert t)=\frac{\exp(v_e^{T}v_t)}{\sum_{e^{\prime}\in E_{KB}}\exp(v^T_{e^{\prime}}v_t)}.
    \label{eq:iv}
\end{equation}

We optimize the NEP model with a cross-entropy loss, but using Eq.~\ref{eq:iv} as is would be computationally expensive as it involves a summation over all entities in the KB.
We address this by replacing $E_{KB}$ in Eq. \ref{eq:iv} with $E^{\ast}$, the union of the positive entity $e$ and $50$
randomly chosen negative entities not in $t$.
Through exploiting the Named Entity Embedding (NEE), our model can represent and thus generate more accurate entities.
The NEE model is not jointly trained with the template components prediction and caption generation heads of \name, but pre-trained offline on 
Wikipedia KB. 

The Wikipedia KB contains a large set of NEs but cannot cover all NEs that could appear in a news article (about $40\%$ are not covered in our datasets). The embedding of a new NE cannot be obtained directly by lookup.
To alleviate this problem, we set the embedding of any missing NE with $v_t$ which is reasonable as we trained the NEP to maximize the correlation between $v_e$ and $v_t$ in Eq.~\ref{eq:iv}.



\subsubsection{Reading Longer Articles}
\label{sec:text}
\citet{Biten_2019_CVPR} use sentence-level features 
obtained by averaging the word features, of a pretrained GloVe \cite{pennington2014glove} model, in the sentence.
While this method can embed the whole article, the averaging makes the feature less informative.
\citet{Tran2020Tell} instead use RoBERTa 
as the text feature extractor, though this has the limitation of exploiting only 512 tokens.

However, processing only the first 512 tokens may ignore important contextual information 
appearing later in the news article.
To alleviate this problem, we propose a \textit{Multi-Span Text Reading} (MSTR) method to read more than $512$ tokens from the article.
MSTR splits the text into overlapping segments of $512$ tokens and pass them to the RoBERTa encoder independently.
The representation of any overlapping token in 2 segments is the element-wise interpolation of their representations.




\section{Experiments}

\begin{table*}[t]
\small
	\centering
	\begin{adjustbox}{max width=\textwidth}
	\begin{tabular}{cc|cccc|cc|cc}
		\cmidrule{3-10}
         & & \multicolumn{4}{c|}{\centering\textbf{\small{General Caption Ceneration}}}
		 & \multicolumn{2}{c|}{\textbf{\small{Named Entities}}} &
        \multicolumn{2}{c}{\textbf{\small{Components}}}\\
		 \cmidrule{3-10}
		 &  & \small{BLEU-4} & \small{ROUGE} & \small{METEOR} & \small{CIDEr} & \small{$P$} & \small{$R$} & \small{$\overline{P}$} & \small{$\overline{R}$} \\
		\midrule
		\multirow{12}{*}[-1cm]{\rotatebox[origin=c]{90}{\good}}
		& SAT~\cite{Kelvin2015} &  0.73 & 11.88 & 4.14 & 12.15 & 8.19 & 7.10 & -- & -- \\
		& Att2in2~\cite{Rennie2017} & 0.76 & 11.58 & 3.90 & 11.58 & -- & -- & -- & -- \\
		& BUTD~\cite{Anderson2018} & 0.71 & 11.06 & 3.74 & 11.02 & -- & -- & -- & -- \\
		& Adaptive Att \cite{Lu2017Adaptive} & 0.51 & 10.94 & 3.59 & 10.55 & -- & -- & -- & -- \\
		& Avg+CtxIns~\cite{Biten_2019_CVPR} & 0.89 & 12.20 & 4.37 & 13.10 & 8.23 & 6.06 & 20.51 & 18.72 \\
		& TBB+AttIns~\cite{Biten_2019_CVPR} & 0.76 & 12.20 & 4.17 & 12.70 & 8.87 & 5.64 & 20.23 & 18.45  \\
		\cmidrule{2-10}
		& VGG+LSTM \cite{Ramisa18} & 0.31 & 6.38 & 1.66 & 1.28 & -- & -- & -- & -- \\
		& VisualNews \cite{Liu2020VisualNewsB} & 5.1 & 19.3 & 8.8 & 43.7 & 19.6 & 17.9 & -- & --\\
		& Tell~\cite{Tran2020Tell} 
		& 5.45 & 20.70 & 9.74 & 48.50 & 21.10 & 17.40 & 69.52 & 63.31 \\
		& Tell (full)~\cite{Tran2020Tell} & 6.05 & 21.40 & 10.30 & 53.80 & 22.20 & 18.70 & 71.55 & 64.93 \\
		 \cmidrule{2-10}
		 & \name (zero-out text) & 1.71 & 13.04 & 5.23 & 9.61 & 4.42 & 3.01 & 18.92 & 16.77 \\
		 & \name (zero-out image) & 4.10 & 17.33 & 8.41 & 38.49 & 18.03 & 15.12 & 48.74 & 46.29 \\
		 & \name (image only) & 1.86 & 13.28 & 5.97 & 10.20 & 4.46 & 3.31 & 19.07 & 17.13  \\
		 & \name (text only) & 5.28 & 19.07 & 9.17 & 50.04 & 20.43 & 18.13 & 49.56 & 46.98  \\ 
		 & \name (auto) & 6.34 & 21.65 & 10.78 & 59.19 & 24.60 & 20.90 & 75.51 & 66.27 \\
		 &  \namens+\neeshort (auto) & 6.73 & 22.68 & 11.18 & 59.50 & 25.87 & 21.63 & 74.42 & 68.53 \\
		 & \namens+MSTR (auto) & 6.45 & 21.99 & 10.83 & 59.65 & 24.75 & 21.61 & 75.57 & \textbf{70.04} \\
		 & \namens+MSTR+\neeshort (auto) & \textbf{6.83} & \textbf{23.05} & \textbf{11.25} & \textbf{61.22} & \textbf{26.87} & \textbf{22.05} & \textbf{75.83} & 
		 68.85 \\
		 \cmidrule{2-10}
		 & \name (oracle) & \textit{7.06} & \textit{24.13} & \textit{11.72} & \textit{69.23} & \textit{28.40} & \textit{23.48} & \textit{92.96} & \textit{87.86} \\
		 & \namens+MSTR+\neeshort (oracle) & \textit{7.36} & \textit{24.25} & \textit{11.98} & \textit{69.76} & \textit{28.59} & \textit{23.68} & \textit{92.46} & \textit{87.55}  \\
		\midrule
		\midrule
		\multirow{8}{*}[-0.5cm]{\rotatebox[origin=c]{90}{\nyt}}
		 & Tell~\cite{Tran2020Tell} 
		 & 5.01 & 19.40 & 9.05 & 40.30 & 20.0 & 18.10 & 67.13 & 62.24 \\
		 & Tell (full)~\cite{Tran2020Tell} & 6.30 & 21.70 & 10.30 & 54.40 & 24.60 & 22.20 & 69.72 & 63.52 \\
		 \cmidrule{2-10}
		 & \name (zero-out text) & 1.42 & 12.66 & 5.08 & 9.33 & 4.23 & 2.89 & 18.87 & 16.53 \\
		 & \name (zero-out image) & 3.88 & 15.64 & 7.76 & 32.01 & 21.15 & 14.84 & 53.71 & 51.29 \\
		 & \name (image only) & 1.50 & 12.58 & 5.68 & 9.93 & 4.49 & 2.88 & 19.40 & 17.12  \\
		 & \name (text only) & 4.95 & 18.47 & 8.54 & 41.27 & 20.52 & 18.48  & 54.89 & 52.31 \\
		 
		 & \name (auto) & 6.39 & 22.38 & 10.75 & 56.54 & 27.35 & 23.73 & 73.37 & 65.79 \\
		 &  \namens+\neeshort (auto) & 6.66 & 22.72 & 10.85 & 59.02 & 26.81 & 23.20 & 73.02 & \textbf{66.54} \\
		 &  \namens+MSTR (auto) & 6.44 & 22.63 & 10.88 & 57.61 & 26.41 & 23.67 & 73.36 & 66.30 \\
		 & \namens+MSTR+\neeshort (auto) & \textbf{6.79} & \textbf{22.80} & \textbf{10.93} & \textbf{59.42} & \textbf{28.63} & \textbf{24.49} & \textbf{73.51} & 65.49 \\
		 \cmidrule{2-10}
		 & \name (oracle) & \textit{7.44} & \textit{24.09} & \textit{11.93} & \textit{65.53} & \textit{28.53} & \textit{26.09} & \textit{90.76} & \textit{87.99} \\
		 & \namens+MSTR+\neeshort (oracle) & \textit{7.68} & \textit{24.09} & \textit{12.09} & \textit{66.15} & \textit{28.79} & \textit{26.35} & \textit{90.07} & \textit{87.92} \\
		\bottomrule
	\end{tabular}
	\end{adjustbox}
	\caption {Results on \good and \nyt. We highlight the \textbf{best} model in bold. Note that we directly use the results reported in~\cite{Tran2020Tell} for the baseline models.\label{tab:results}}
	\vspace{-0.3cm}
\end{table*}

We evaluate \name on two large-scale publicly available news captioning datasets: 
\good \cite{Biten_2019_CVPR} 
and 
\nyt 
\cite{Tran2020Tell}
both collected using The New York Times public API\footnote{https://developer.nytimes.com/apis}, with the latter being larger and containing longer articles. 
We follow the evaluation protocols defined by the authors of each dataset and used by previous works with $421$K training, $18$K validation, and $23$K test captions for \good and $763$K training, $8$K validation and $22$K test captions for \nyt.
We provide further details about the datasets in the supplementary material.


\subsection{Methods \& Metrics}

We implement \name as a Transfomer-based 
encoder-decoder architecture similar to Tell but with our proposed template guidance.
We introduce \namens+\neeshort as \name with enriched named entity embeddings (Section~\ref{sec:nee}), and \namens+MSTR as \name with multi-span text reading technique (Section~\ref{sec:text}).
To evaluate how \name exploits template guidance, we introduce the \name (oracle) and \namens+MSTR+\neeshort (oracle) variants, where ground truth template components are provided through $\bm{\alpha}$. 
We evaluate if our model exploits both the text and image input in two ways. 
We first report results of our multimodal model where at test time we zero-out text features (i.e. $\mathbf{X}^T$ and $\mathbf{X}^E$ are set to all zero vectors) \name (zero-out text) or image features 
\name (zero-out image). 
We also train single-modality models with only an image encoder (\name image only) or a text encoder (\name text only).


We compare against two types of baselines.
(i) Two-step generation methods: that are based on conventional image captioning models \cite{Kelvin2015,Rennie2017,Anderson2018,Lu2017Adaptive,Biten_2019_CVPR} to first generate captions with
placeholders and then insert named entities into these placeholders.
(ii) End-to-end models: VGG+LSTM~\cite{Ramisa18}, VisualNews~\cite{Liu2020VisualNewsB} that uses ResNet as image encoder,
BERT article encoder and bi-LSTM as decoder, and Tell, with two variants: 
(a) Tell, which uses RoBERTa and ResNet-152 as the encoders and Transformer as the decoder, it is equivalent to \name without template guidance as they use the same encoders and training settings.
(b) Tell (full), which includes two additional visual encoders: YOLOv3 and MTCNN, and Location-Aware and Weighted RoBERTa for text encoding.

For the general caption generation quality evaluation, we use the BLEU-4 \cite{papineni2002}, ROUGE \cite{lin2004rouge}, METEOR \cite{denkowski2014meteor} and CIDEr \cite{Vedantam15cider} metrics.
We also use named entity precision/recall
to evaluate the named entity generation quality.
To better understand how well the generated captions follow the ground truth templates, we calculate precision and recall for the five components \textit{who}, \textit{when}, \textit{where}, \textit{context} and \textit{misc} and use the averaged precision and recall\footnote{Per-component results are provided in Table 4 of the supplementary material.} as the final metric.

\subsection{Implementation and Training details}

Following \citet{Tran2020Tell}, we set the hidden size of the input features $d_I=2048$, $d_T=1024$ and $d_E=300$ and the number of heads $H=16$.
We use the Adam optimizer~\cite{kingma2015} with $\beta_1=0.9$, $\beta_2=0.98$, $\epsilon=10^{-6}$.
The number of tokens in the vocabulary $K=50264$ and $d^{Wiki}=300$.
We limit the text length in MSTR to 1,000 tokens 
as preliminary studies have shown similar performance with longer text input but at the expense of significant increased training time (Tab. 6 in supplementary). 
In practice, for an article longer than 512 tokens, we read two overlapping text segments of 512 tokens, one starting from the beginning and another from the end and thus can have $[24-511]$ overlapping tokens.
The components prediction head in Fig.~\ref{fig:arch} is a linear layer followed by an output layer of $1024$ dimensions.

The training pipeline uses PyTorch~\cite{paszke2017automatic} and
the AllenNLP framework~\cite{gardner2018allennlp}. The RoBERTa model and
dynamic convolution code are adapted from fairseq~\cite{ott2019fairseq}.
We use a maximum batch size of $16$ and training is stopped after the model has seen 6.6 million examples, corresponding to $16$ epochs on \good and $9$ epochs on \nyt. 
Training is done with mixed precision to reduce the memory footprint and allow our full model to be trained on a single V-100 GPU for 4 to 6 days on both datasets.

\subsection{Evaluation}



\begin{figure*}[t]
    \centering
    \small
    \includegraphics[width=\textwidth]{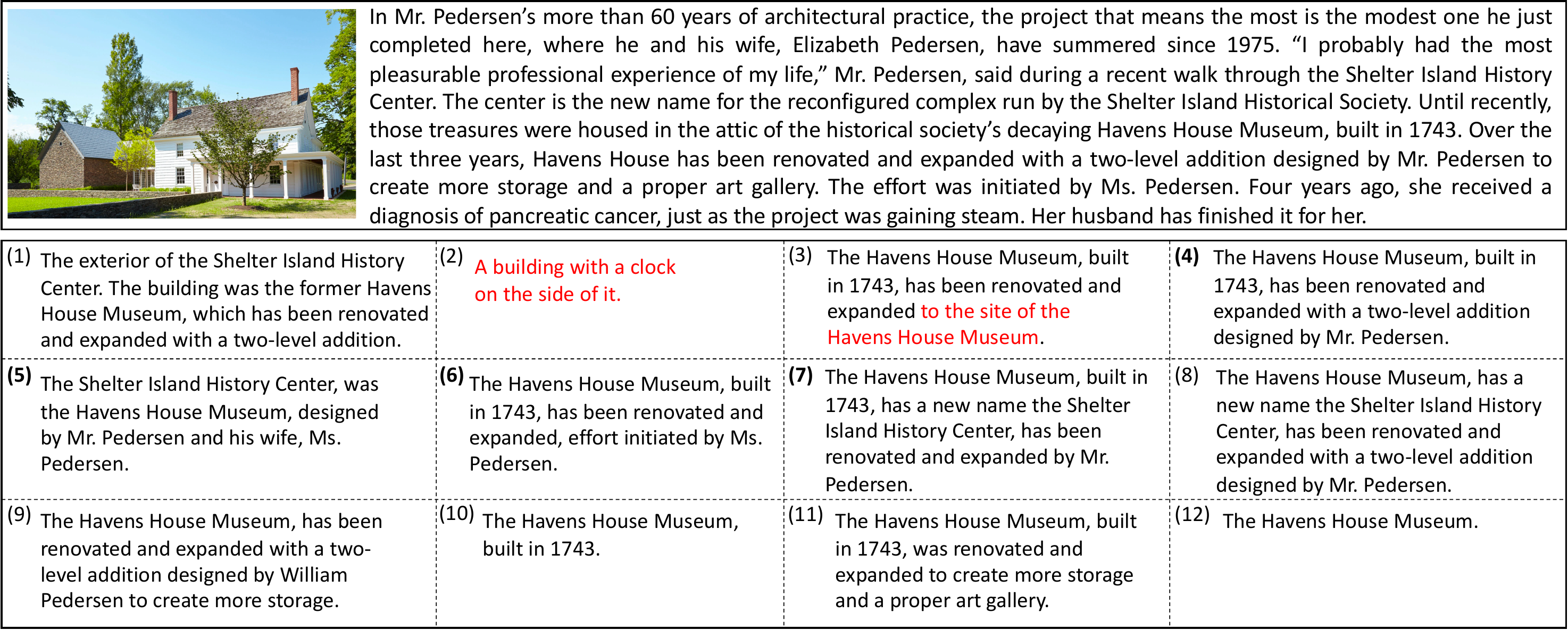}
    \caption{An example of news caption generation. The captions are generated by: (1) human (ground truth caption). (2) conventional image captioning model SAT. (3) Tell. (4)  \name. (5)  \namens+\neeshort. (6) \namens+MSTR. 
    \namens+MSTR+\neeshort (7) auto, (8) oracle, with template (9) \textit{who} + \textit{context}, (10) \textit{who} + \textit{when}, (11) \textit{who} + \textit{when} + \textit{context}, and (12) \textit{who}.
     For the generated captions, we highlight wrong statements in {\color{red}red}.
    }
    \label{fig:qualitative}
\vspace{-0.3cm}
\end{figure*}

\subsubsection{General Caption Generation}
We first discuss the results with the general caption generation metrics BLEU-4, ROUGE, METEOR and CIDEr reported in Table~\ref{tab:results}.
We report the mean values of three runs, and the maximum standard deviations of our variants on BLEU, ROUGE, METEOR, CIDEr are 0.013, 0.019, 0.016 and 0.069, which shows the stability of our results and that our method improvements are notable.
For the \good dataset, \name (auto) provides an improvement of $0.89$, $0.95$, $1.04$, $10.69$ points over Tell on the four metrics respectively, while the full model \namens+MSTR+\neeshort (auto) has an even bigger improvement of $1.38$, $2.35$, $1.51$, $12.72$.
The improvement is especially impressive for the CIDEr score.
\name performs much better than all the two-step captioning methods (first group of results) and VGG+LSTM.
For the \nyt dataset, we compare our models only to Tell since other models perform much worse.  
Here, our full model 
achieves $6.79$, $22.80$, $10.93$ and $59.42$ 
with $1.78$, $3.40$, $1.88$, $19.12$ points improvement over Tell.
Our \namens+MSTR+\neeshort (auto) outperforms Tell (full) which exploits additional visual features on both datasets .
This demonstrates the effectiveness of our model in generating good captions.
By providing the oracle  $\bm{\alpha}$, the \namens+MSTR+\neeshort (oracle) can achieve even higher performance on almost all metrics, showing the value of our template guidance process. 

From the single modality evaluation, we observe that models that exploit the text only (\name (zero-out image) and \name (text only)) perform better than those relying on the image only (\name (zero-out text) and \name (image only)) but all have lower performance than multimodal models, confirming that both modalities are important for news image captioning.

\subsubsection{Named Entity Generation}
One of the main objectives of news captioning is to generate captions with accurate named entities.
As shown in Tab.~\ref{tab:results}, compared to Tell, \namens+MSTR+\neeshort (auto) increases the named entity precision and recall scores by $5.77\%$ 
and $4.65\%$ 
on \good, and $8.63\%$ 
and $6.39\%$ 
on \nyt.
The oracle versions of our models attain even higher performances.

\subsubsection{Template Components Evaluation}
The average precision and recall of the template components, reported in the two rightmost columns of Tab.~\ref{tab:results},
of \namens+MSTR+\neeshort (auto)  increases by 
$6.3\%$
and 
$5.5\%$ 
on \good dataset and 
$6.4\%$ 
and 
$3.3\%$ 
on \nyt dataset compared to Tell.
By providing the oracle $\bm{\alpha}$, 
even better results are obtained, demonstrating that our model can exploit template guidance.


\subsubsection{Qualitative \& Human Evaluation}
In Figure~\ref{fig:qualitative}  
we show the image, article (shortened for visualization) and the captions generated by a conventional image captioning model SAT~\cite{Kelvin2015}, Tell~\cite{Tran2020Tell} and different \name variants.
The captions generated by all \name variants are meaningful and closer to the ground truth than the baselines.
Interestingly, most captions generated by \name variants include people's names, e.g. Mr. or Ms. \textit{Pedersen}
in addition to the building names
probably because people's names are the most common type for the component \textit{who} in the datasets (see Tab. 1 of the supplementary material).
As MSTR can read longer text than Tell, 
\namens+MSTR can exploit 
the end of the article and generates the text span \textit{effort initiated by Ms. Pedersen}.
The caption generated by \namens+MSTR+\neeshort has all the key factors in the ground truth caption (\textit{the Havens House Museum}, \textit{the Shelter Island History Center}, \textit{been renovated and expanded}) 
demonstrating the strengths of our model.
The captions generated using the oracle $\bm{\alpha}$ (8) as well as some other manually defined $\bm{\alpha}$ (9-12)
 illustrate the benefits and flexibility
of our ``template components'' modeling, showing
how the caption generation process can be controlled by the template guidance in \name.



Finally, we conducted a human evaluation through crowd-sourcing on Amazon Mechanical Turk on 200 random image-article pairs sampled from the test set of the NYT800K dataset. 
For each image-article pair, three different raters
were requested to rate the ground truth caption, the caption generated by Tell, and captions generated by 4
variants
of our model, on a 4 point scale.
Raters were asked to evaluate separately  how well the caption was describing the \textit{image}, how relevant it was to the \textit{article}, and how easy to understand the \textit{sentence} was.
We report the average of the three ratings in Tab.~\ref{tab:human}, showing that all variants of our model produce captions that are rated better than Tell and closer to the ground truth captions ratings on the three aspects.
The groundtruth captions have the highest sentence quality score but can have lower score for image and article relatedness as journalists sometimes do not follow guidelines and can write a caption describing the image independently of the article context or on the contrary being more related to the article than the image content. 
Details on the annotation instructions and results are given in the supplementary material.

\begin{table}[!t]
\centering
\small
\begin{tabular}{cccccccl}
\toprule
\textbf{Model} & image & article & sentence \\
\midrule
Ground Truth & 2.96 & 2.86 & 3.08 \\ 
\midrule
Tell & 2.80 & 2.80 & 2.92 \\
\midrule
\name & 2.87 & 2.86 & 2.97 \\
\namens+\neeshort & 2.88 & \textbf{2.92} & \textbf{2.99} \\
\namens+MSTR & \textbf{2.89} & 2.86 & 2.98 \\
\namens+MSTR+\neeshort & 2.86 & 2.88 & \textbf{2.99} \\
\bottomrule
\end{tabular}

\caption{\textbf{Human evaluation on the generated captions.} We highlight the \textbf{best} model in bold.}
\label{tab:human}
\vspace{-0.35cm}
\end{table}

\section{Conclusion}
News image captioning is a challenging task as it requires exploiting both image and text content to produce rich and well structured captions including relevant named entities and information gathered from the whole article.  
In this work, we presented \fullname, aiming to solve the news image captioning task by integrating domain specific knowledge in both the representation and caption generation process. On the representation side, we introduced two techniques: named entity embedding (\neeshortns) and multi-span text reading (MSTR).
Our decoding process explicitly integrates the key components a journalist would seek to describe to  improve the caption generation quality.
Our method obtains remarkable gains on both \good and \nyt 
datasets relative to the state-of-the-art.



\section{Ethical Considerations and Broader Impact}
Our model is a multi-modality extension of the general image captioning methods.
It can further be applied to other applications, including but not limited to, multi-modality machine translation, summarization, etc.
By modeling the template components of the captions, our research could be used to explore the underlying structure of each task, improving understanding of the generation decisions or providing explanations.
The potential risks of news article image captioning is the generation bias, i.e., the model might tend to use the named entities that have high frequencies.
We thus suggest that people use our model as a recommendation for generating captions, 
people could thus modify the generated captions and control for potential bias.
We would also encourage further work to understand the biases and limitations of the datasets used in this paper, including tools to analyze gender bias and other limitations. 

\section*{Acknowledgments}
We thank Mahdi Abavisani, Shengli Hu, and Di Lu for the fruitful discussions during the development of the method, and all the reviewers for their detailed questions, clarification requests, and suggestions on the paper.

\bibliography{eacl2021}
\bibliographystyle{acl_natbib}

\appendix

\section{Appendices}

The appendices provide more information about the two datasets \good and \nyt, template statistics and prediction results, implementation and training details, model difference between Tell and \name, details on the human evaluation and ablation evaluations of another sequence length efficient Transfomer - Longformer as well as different sequence length for MSTR.


\subsection{Datasets}
We use two datasets GoodNews \cite{Biten_2019_CVPR} and NYTimes800k \cite{Tran2020Tell}.
Both datasets are collected by using The New York Times public API\footnote{https://developer.nytimes.com/apis}
For GoodNews dataset, since only the articles, captions, and
image URLs are publicly released, the images need to be
downloaded from the original source. 
Out of the 466K image URLs provided by \citet{Biten_2019_CVPR}, we were able to download 463K images, the remaining are broken links.
We use the same train, validation and test splits provided as \citet{Biten_2019_CVPR}. There are $421$K training, $18$K validation, and $23$K test captions.

NYTimes800k dataset is $70\%$ larger and more complete dataset of New York Times articles, images, and captions. 
The number of train, validation and test sets are $763$K, $8$K and $22$K respectively.
Tab.~\ref{tab:dataset} presents a detailed comparison between GoodNews and NYTimes800k in terms of articles and captions length, and captions composition.

\begin{table}[t]
\small
    \centering
    \begin{tabular}{ccc}
    \toprule
    & GoodNews & NYTimes800k \\\midrule
    \# of articles & 257033 & 444914 \\
    \# of images & 462642 & 792971 \\
    Average article length & 653 & 892 \\
    Average caption length & 18 & 18 \\\hline
    \% of caption words that are & & \\
    nouns & 16\% & 16\% \\
    pronouns & 1\% & 1\% \\
    proper nouns & 23\% & 22\% \\
    verbs & 9\% & 9\% \\
    adjectives & 4\% & 4\% \\
    named entities & 27\% & 26\% \\ \hline
    \% of captions with & & \\
    named entities & 97\% & 96\% \\
    people’s names & 68\% & 68\% \\
    \bottomrule
    \end{tabular}
    \caption{Summary of news captioning datasets.}
    \label{tab:dataset}
\end{table}

\begin{table}[t]
    \centering
    \small
    \begin{tabular}{cccccc}
    \toprule
      Dataset & average len  & \% $> 512$ & \% $> 1000$  \\
      \midrule
      \good & 653  & 49.7\% & 18.2\% \\
      \nyt & 892  & 54.85\% & 21.92\% \\
      \bottomrule
    \end{tabular}
    \caption{Article length statistics for the \good and \nyt dataset.}
    \label{tab:len_stat}
    \vspace{-0.5cm}
\end{table}

We also show the article length statistics in Tab.~\ref{tab:len_stat}.
With approximate $50\%$ of the training articles having more than 512 tokens, MSTR technique is necessary to deal with this problem.

\subsection{Template statistics and prediction results}

\begin{table*}[t]
    \centering
    \small
    \begin{adjustbox}{max width=\textwidth}
    \begin{tabular}{ccccccccccccccccc}
    \toprule
       template  & 1 & 2 & 3 & 4 & 5 & 6 & 7 & 8 & 9 & 10 & 11 & 12 & 13 & 14 & 15 & avg \\\midrule
        \% & 15.2 & 4.4 & 4.2 & 3.5 & 2.8 & 2.6 & 13.1 & 12.7 & 7.7 & 7.3 & 6.8 & 5.3 & 5.1 & 2.4 & 2.2 & -- \\
        who & $\times$ & $\times$ & $\times$ & $\times$ & $\times$ & $\times$ & $\times$ & $\times$ & $\times$ & $\times$ & $\times$ & $\times$ & $\times$ & -- & -- & -- \\
        when & $\times$ & -- & -- & $\times$ & -- & $\times$ & -- & -- & $\times$ & -- & $\times$ & -- & $\times$ & -- & -- & -- \\
        where & $\times$ & -- & $\times$ & $\times$ & -- & -- & -- & $\times$ & -- & -- & $\times$ & $\times$ & -- & $\times$ & -- & -- \\
        misc & $\times$ & -- & -- & -- & $\times$ & -- & -- & -- & -- & $\times$ & $\times$ & $\times$ & $\times$ & -- & -- & -- \\
        context & -- & -- & -- & -- & -- & -- & $\times$ & $\times$ & $\times$ & $\times$ & $\times$ & $\times$ & $\times$ & $\times$ & $\times$ & -- 
    \end{tabular}
    \end{adjustbox}
    \caption{Template class definition and its relation with different components. Note there are in total $2^5$ template class but we show the ones with over $2\%$ of samples which accounts for $96.2\%$ of the training data.}
    \label{tab:template_stat}
         \vspace{-0.3cm}
\end{table*}

\begin{table*}[t]
\small
	\centering
	\begin{adjustbox}{max width=\textwidth}
	\begin{tabular}{cc|cccccccccccc}
		\cmidrule{3-14}
         & &
        \multicolumn{2}{c}{\textbf{\small{Average}}} & \multicolumn{2}{c}{\textbf{\small{who}}} & \multicolumn{2}{c}{\textbf{\small{when}}} & \multicolumn{2}{c}{\textbf{\small{where}}} & \multicolumn{2}{c}{\textbf{\small{misc}}} & \multicolumn{2}{c}{\textbf{\small{context}}} \\
		 \cmidrule{3-14}
		  & & \small{$\overline{P}$} & \small{$\overline{R}$}  & P & R & P & R & P & R & P & R & P & R  \\
		\midrule
		\multirow{6}{*}[-1cm]{\rotatebox[origin=c]{90}{\good}}
		
		& Tell~\cite{Tran2020Tell} & 69.52 & 63.31 & 90.44 & 83.48 & 59.25 & 58.78 & 62.90 & 66.67 & 51.59 & 42.50 & 83.43 & 65.13 \\
		& Tell (full)~\cite{Tran2020Tell} & 71.55 & 64.93 & 91.49 & 85.89 & 63.04 & 60.69 & 65.45 & 66.90 & 53.27 & 46.00 & 84.50 & 65.19 \\
		 \cmidrule{2-14}
		 & \name (zero-out text) &  18.92 & 16.77 & 23.15 & 21.28 & 16.96 & 15.12 & 18.63 & 17.86 & 13.14 & 11.58 & 22.72 & 18.01 \\
		 & \name (zero-out image) & 48.74 & 46.29 & 63.09 & 60.21 & 39.82 & 37.40 & 44.96 & 42.60 & 25.50 & 23.03 & 70.33 & 68.21 \\
		 & \name (image only) &  19.07 & 17.13 & 23.38 & 21.61 & 17.15 & 15.52 & 18.70 & 17.92 & 13.31 & 11.72 & 22.81 & 18.88 \\
		 & \name (text only) & 49.56 & 46.98 & 63.89 & 60.89 & 40.73 & 38.11 & 45.81 & 43.25 & 26.20 & 23.73 & 71.17 & 68.92 \\
		 & \name (Longformer) & 74.07 & 58.86 & 94.46 & 87.45 & 63.29 & 26.66 & 71.45 & 62.66 & 54.67 & 45.30 & 86.46 & 72.25 \\
		 & \name (auto) & 75.51 & 66.27 & 94.77 & 86.59 & 66.15 & 64.19 & 71.21 & 67.79 & \textbf{58.92} & 46.12 & 86.52 & 66.65 \\
		 &  \namens+\neeshort (auto) & 74.42 & 68.53 & 94.64 & 88.65 & 64.66 & 64.26 & 70.54 & 68.22 & 55.93 & 48.65 & 86.34 & 72.88 \\
		 & \namens+MSTR (auto) & 75.57 & \textbf{70.04} & 94.32 & \textbf{91.00} & \textbf{68.75} & \textbf{66.91} & 71.50 & \textbf{70.16} & 56.81 & \textbf{49.07} & 86.49 & \textbf{73.07} \\
		 & \namens+MSTR+\neeshort (auto) &  \textbf{75.83} & 
		 68.85 & \textbf{95.75} & 90.01 & 66.72 & 64.45 & \textbf{72.19} & 69.38 & 57.33 & 48.48 & \textbf{87.19} & 71.96 \\
		 \cmidrule{2-14}
		 & \name (oracle) &  \textit{92.69} & \textit{87.86} & \textit{95.07} & \textit{88.21} & \textit{97.09} & \textit{95.10} & \textit{88.50} & \textit{84.02} & \textit{84.07} & \textit{78.01} & \textit{98.75} & \textit{93.97} \\
		 & \namens+MSTR+\neeshort (oracle) &  \textit{92.46} & \textit{87.55} & \textit{95.07} & \textit{88.79} & \textit{97.00} & \textit{93.86} & \textit{88.09} & \textit{83.79} & \textit{83.43} & \textit{75.81} & \textit{98.70} & \textit{95.50}  \\
		\midrule
		\midrule
		\multirow{6}{*}[-0.5cm]{\rotatebox[origin=c]{90}{\nyt}}
		 & Tell~\cite{Tran2020Tell} & 67.13 & 62.24 & 86.44 & 79.65 & 57.45 & 63.93 & 61.08 & 72.19 & 46.30 & 36.99 & 84.39 & 58.44 \\
		 & Tell (full)~\cite{Tran2020Tell} & 69.72 & 63.52 & 88.91 & 82.92 & 61.30 & 65.83 & 63.97 & 73.52 & 49.40 & 39.34 & 85.07 & 56.01 \\
		 \cmidrule{2-14}
		 & \name (zero-out text) & 18.87 & 16.53 & 22.96 & 20.96 & 16.52 & 14.88 & 18.01 & 17.21 & 13.02 & 11.34 & 23.84 & 18.26 \\
		 & \name (zero-out image) & 53.71 & 51.29 & 79.74 & 68.10 & 41.83 & 45.91 & 51.07 & 48.82 & 26.89 & 30.19 & 69.02 & 63.43 \\
		 & \name (image only) & 19.40 & 17.12 & 23.51 & 21.55 & 16.98 & 15.51 & 18.54 & 17.69 & 13.61 & 11.82 & 24.36 & 19.03 \\
		 & \name (text only) & 54.89 & 52.31 & 80.96 & 69.08 & 44.02 & 47.02 & 52.39 & 49.75 & 28.15 & 31.23 & 68.93 & 64.47 \\
		 & \name (Longformer) & 68.93 & 56.67 & 88.24 & 83.72 & 60.59 & 28.63 & 65.10 & 65.73 & 46.64 & 38.87 & 84.08 & 66.40 \\
		 & \name (auto) & 73.37 & 65.79 & 92.89 & 85.61 & \textbf{64.62} & \textbf{66.90} & 69.69 & 72.22 & \textbf{53.25} & 41.53 & 86.38 & 62.69 \\
		 &  \namens+\neeshort (auto) & 73.02 & \textbf{66.54} & \textbf{93.54} & \textbf{86.45} & 63.54 & 64.44 & 68.43 & \textbf{74.59} & 52.81 & \textbf{44.37} & 86.80 & 62.86 \\
		 &  \namens+MSTR (auto) & 73.36 & 66.30 & 93.21 & 85.97 & 64.59 & 66.60 & 70.32 & 73.25 & 51.67 & 42.62 & 87.02 & \textbf{63.08} \\
		 & \namens+MSTR+\neeshort (auto) &  \textbf{73.51} & 65.49 & 93.10 & 86.44 & 64.09 & 65.05 & \textbf{70.40} & 74.26 & 52.82 & 41.05 & \textbf{87.15} & 60.66 \\
		 \cmidrule{2-14}
		 & \name (oracle) & \textit{90.76} & \textit{87.99} & \textit{93.35} & \textit{87.63} & \textit{95.42} & \textit{94.04} & \textit{87.59} & \textit{86.57} & \textit{78.58} & \textit{76.08} & \textit{98.86} & \textit{95.65} \\
		 & \namens+MSTR+\neeshort (oracle) &  \textit{90.07} & \textit{87.92} & \textit{92.88} & \textit{88.16} & \textit{94.70} & \textit{93.59} & \textit{86.86} & \textit{87.28} & \textit{77.26} & \textit{75.03} & \textit{98.67} & \textit{95.55} \\
		\bottomrule
	\end{tabular}
	\end{adjustbox}
	\caption {Precision and Recall results of each template component prediction on \good and \nyt. We highlight the \textbf{best} model in bold.\label{tab:results}}
\end{table*}

We show the composition in terms of components and the percentage of the template classes of the whole GoodNews dataset in Tab.~\ref{tab:template_stat}.

We also report in Tab.~\ref{tab:results} detailed template components precision and recall scores for different variants of our model and the Tell baseline on the two datasets.

\subsection{Implementation and Training details}

Following \citet{Tran2020Tell}, we set the hidden size of the input features $d_I=2048$, $d_T=1024$ and $d_E=300$ and the number of heads $H=16$.
We use the Adam optimizer~\cite{kingma2015} with $\beta_1=0.9$, $\beta_2=0.98$, $\epsilon=10^{-6}$.
The number of tokens in the vocabulary $K=50264$ and $d^{Wiki}=300$.
We use a maximum batch size of $16$ and training is stopped after the model has seen 6.6 million examples, corresponding to $16$ epochs on \good and $9$ epochs on \nyt. 
The components prediction head in Fig. 2 of the main paper is a Linear layer followed by an output layer with hidden states dimension equal to $1024$.
The training pipeline is written in PyTorch~\cite{paszke2017automatic} using
the AllenNLP framework~\cite{gardner2018allennlp}. The RoBERTa model and
dynamic convolution code are adapted from fairseq~\cite{ott2019fairseq}.
Training is done with mixed precision to reduce the memory footprint and allow our full model to be trained on a single GPU.
The models take 4 to 6 days to train on one V-100 GPU on both datasets.

\subsection{Model Difference Between Tell and \name}

As shown in Tab.~\ref{tab:tell_vs_tgnc}, our model shares some components with the baseline model Tell~\cite{Tran2020Tell}.
\name and Tell both use an image and text encoder and a Transformer decoder.
However, \name applies template guidance to model the journalistic guidelines for caption generation.

\begin{table*}[t]
    \centering
    \small
    \begin{adjustbox}{max width=\textwidth}
    \begin{tabular}{cccccccccc}
    \toprule
         & image & text & template guidance & faces & objects  & weighted RoBERTa & location aware & decoder & \# of parameters \\\midrule
         Tell & $\times$ & $\times$  &  -- & -- & -- & -- & -- & Transformer & 125M\\
         Tell (full) & $\times$ & $\times$ &  -- & $\times$ & $\times$ & $\times$ & $\times$ & Transformer & 200M \\
         \name & $\times$ & $\times$  &  $\times$ & -- & -- & -- & -- & Transformer & 205M \\
         \bottomrule
    \end{tabular}
    \end{adjustbox}
    \caption{The difference between \name and Tell \cite{Tran2020Tell}. Tell can be regarded as a variant of \name without template guidance. $\times$: having this technique. --: not having this one.}
    \label{tab:tell_vs_tgnc}
\end{table*}

\subsection{Human evaluation}

We have conducted a human evaluation of 200 article-image pairs. Below the article and the image, we displayed either the ground truth caption or a caption generated by Tell or one of our model variant.
We ask the annotators to rate each caption as follows:
\begin{itemize}
    \item How well does the caption describe the IMAGE? Regardless of how fluent it is.
    \begin{itemize}
        \item 1 = Very bad (Does not describe the image)
        \item 2 = Somewhat bad (Describes the image, but contradictory to or missing key information from the image) 
        \item 3 = Somewhat good (Describes the image, no contradictions but missing key information from the image) 
        \item 4 = Very good (Describes the image, no contradictions and contains the key information from the image)
    \end{itemize}
    \item How well does the caption summarize the ARTICLE? Regardless of how fluent it is.
    \begin{itemize}
        \item 1 = Very bad (Not relevant to the topic)
        \item 2 = Somewhat bad (Covers the right topic, but contradicting the article or missing key facts)
        \item 3 = Somewhat good (Covers the right topic, no contradictions with the article, but missing key facts)
        \item 4 = Very good (Covers the right topic, no contradictions with the article, and contains the key facts)
    \end{itemize}
    \item How easy or hard is it to understand the SENTENCE? Regardless of how well it describes the image or article.
    \begin{itemize}
        \item 1 = Very hard or doesn’t make sense
        \item 2 = Somewhat hard
        \item 3 = Somewhat easy
        \item 4 = Very easy to understand
    \end{itemize}
\end{itemize}

\begin{figure}[t]
         \includegraphics[width=\columnwidth]{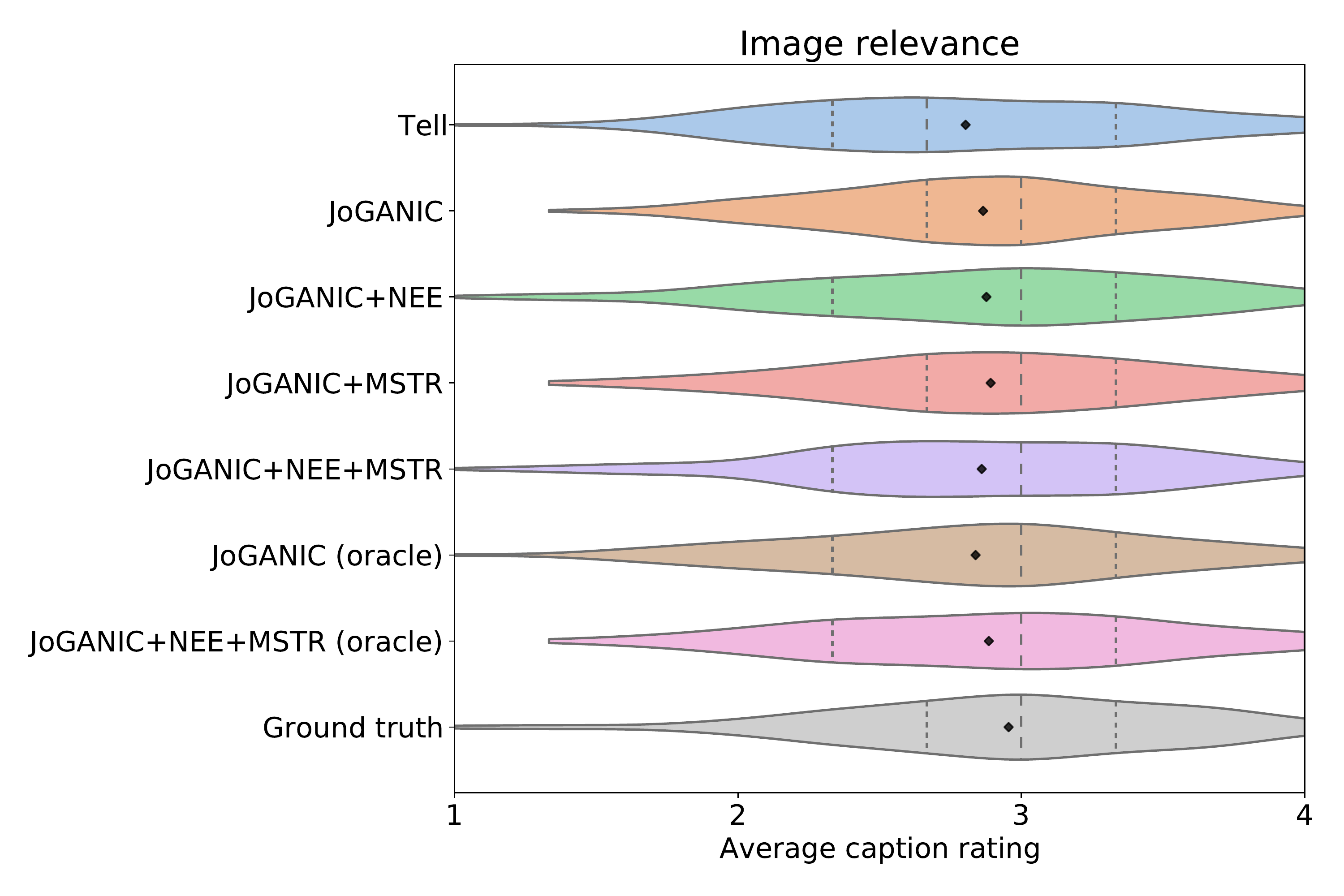}
         \caption{Image relevance ratings distributions.}
         \label{fig:humaneval_img}
         \vspace{-0.3cm}
\end{figure}

\begin{figure}[t]
         \includegraphics[width=\columnwidth]{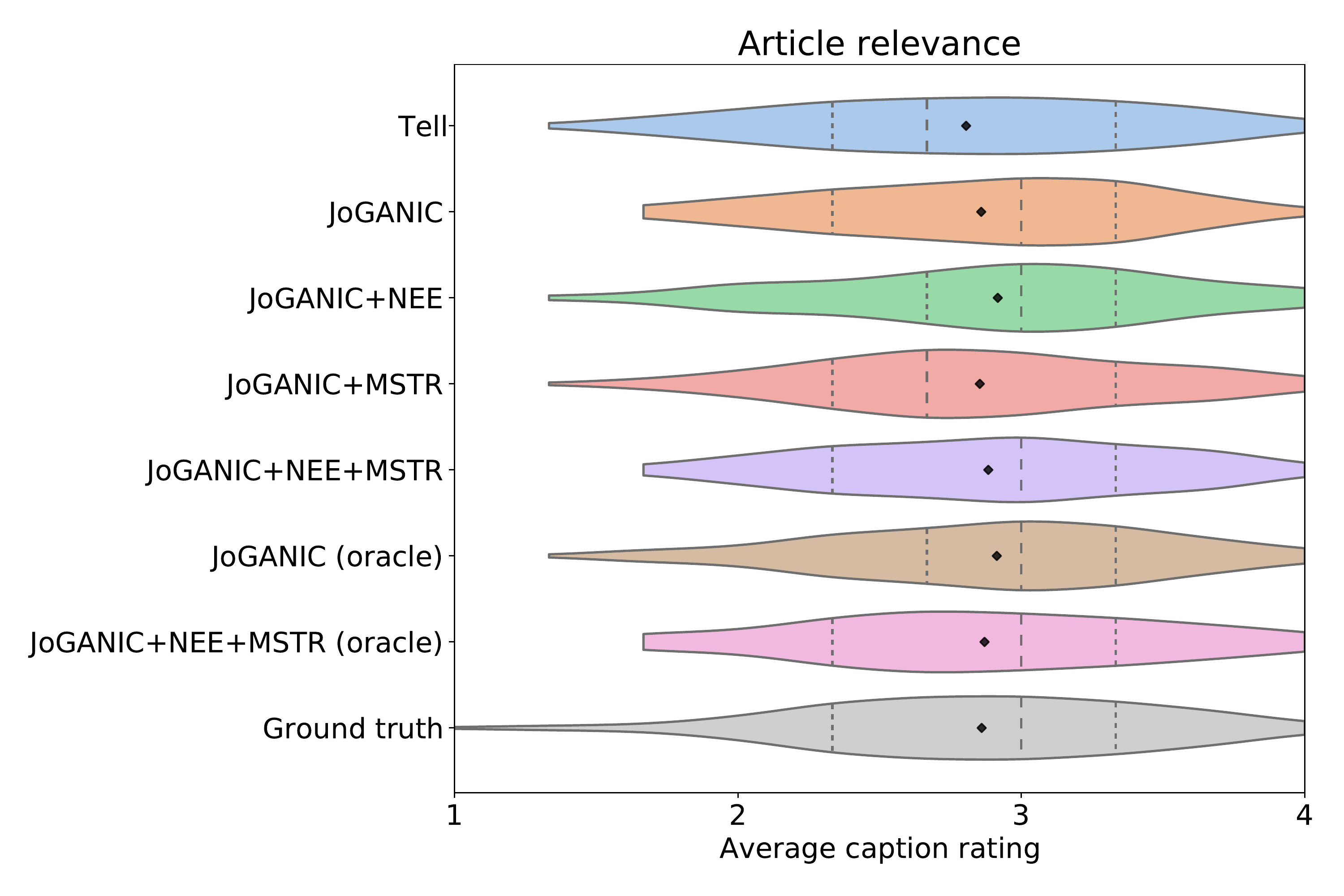}
         \caption{Article relevance ratings distributions.}
         \label{fig:humaneval_art}
         \vspace{-0.3cm}
\end{figure}
     
\begin{figure}[t]
         \includegraphics[width=\columnwidth]{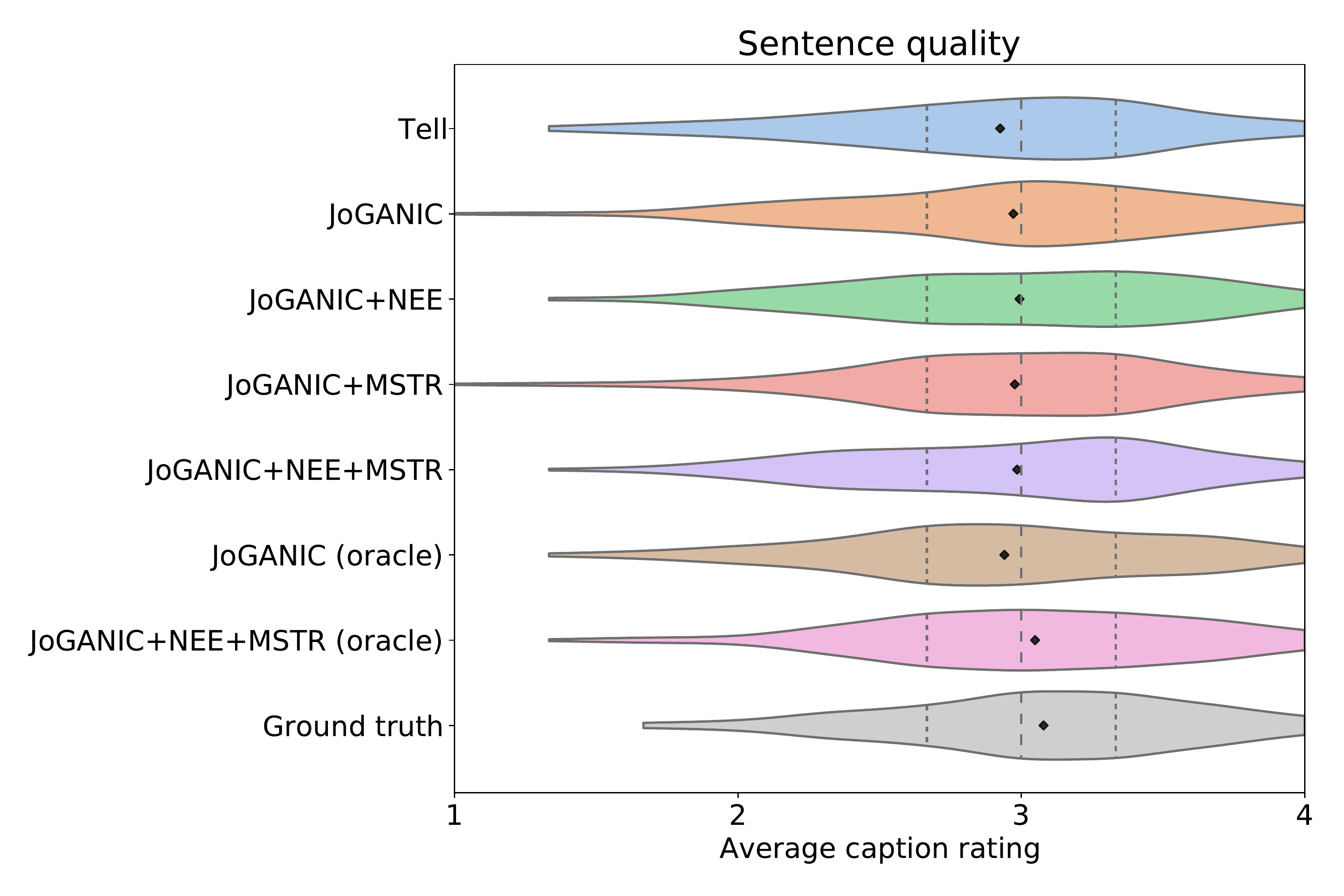}
         \caption{Sentence quality ratings distributions.}
         \label{fig:humaneval_sen}
         \vspace{-0.3cm}
\end{figure}

Each image-article pair is shown to three different annotators, thus each caption is rated three times. We average the rating for each caption, and then plot the image relevance, article relevance and sentence quality ratings statistics as violin plots in Figure~\ref{fig:humaneval_img}, Figure~\ref{fig:humaneval_art} and Figure~\ref{fig:humaneval_sen}, respectively. 
In each of these plots, the median is reported as a large dashed line, the first and third quartile as thinner dashed lines and the mean score as the black diamond.
The varying height of each violin represent the number of samples having the corresponding rating.
We can observe that all distributions are somewhat similar, but the Tell model is generally produces the lowest rated captions. The basic JoGANIC is a bit better, while more advanced variations of our model produce captions that are rated higher and really similarly to the ground truth captions.


\subsection{Ablation Study}
\begin{table*}[t]
\small
	\centering
	\begin{adjustbox}{max width=\textwidth}
	\begin{tabular}{cc|cccc|cc|c}
		\cmidrule{3-9}
         & & \multicolumn{4}{c|}{\centering\textbf{\small{General Caption Ceneration}}}
		 & \multicolumn{2}{c|}{\textbf{\small{Named Entities}}} 
		 & \textbf{\small{Training Time}}
        \\
		 \cmidrule{3-9}
		 &  & \small{BLEU-4} & \small{ROUGE} & \small{METEOR} & \small{CIDEr} & \small{$P$} & \small{$R$} & \small{$h$/epoch}  \\
		\midrule
		\multirow{6}{*}{\good}
		 & \name (Longformer) & 5.69 & 21.08 & 9.97 & 52.04 & 24.63 & 19.63 & \textbf{0.91} \\
		 & \name (RoBERTa) & 6.34 & 21.65 & 10.78 & 59.19 & 24.60 & 20.90 & 1.02 \\
		 & \name (RoBERTa+MSTR 800) & 6.38 & 21.72 & 10.80 & 59.33 & 24.63 & 21.22 & 1.23 \\
		 & \name (RoBERTa+MSTR 1000) & \textbf{6.45} & \textbf{21.99} & 10.83 & 59.65 & \textbf{24.75} & 21.61 & 1.41 \\
		 & \name (RoBERTa+MSTR 1200) & 6.44 & 21.98 & \textbf{10.85} & 59.66 & 24.74 & \textbf{21.63} & 1.58 \\
		 & \name (RoBERTa+MSTR 1400) & 6.45 & 21.96 & 10.80 & \textbf{59.67} & 24.74 & 21.60 & 1.83 \\
		\midrule
		\midrule
		\multirow{6}{*}{{\nyt}}
		 & \name (Longformer) & 5.72 & 19.55 & 9.87 & 41.66 & 22.89 & 18.09 & \textbf{0.94} \\
		 & \name (RoBERTa) & 6.39 & 22.38 & 10.75 & 56.54 & \textbf{27.35} & \textbf{23.73} & 1.09  \\
		 & \name (RoBERTa+MSTR 800) & 6.41 & 22.40 & 10.79 & 56.92 & 27.01 & 23.70 & 1.26 \\
		 & \name (RoBERTa+MSTR 1000) & \textbf{6.44} & 22.63 & \textbf{10.88} & \textbf{57.61} & 26.41 & 23.67 & 1.47 \\
		 & \name (RoBERTa+MSTR 1200) & 6.42 & \textbf{22.64} & 10.83 & 57.59 & 26.43 & 23.61 & 1.69 \\
		 & \name (RoBERTa+MSTR 1400) & 6.42 & 22.62 & 10.81 & 57.60 & 26.40 & 23.58 & 1.88 \\
		\bottomrule
	\end{tabular}
	\end{adjustbox}
	\caption {Results on \good and \nyt. We highlight the \textbf{best} model in bold.
	Note that we report the mean values of three runs, and the maximum standard derivations of our variants on BLEU, ROUGE, METEOR, CIDEr are 0.013, 0.019, 0.016 and 0.069, which shows the stability of our results and that our method improvements are notable.
	\label{tab:ablation_results}}
\end{table*}
In addition to the \textit{Multi-Span Text Reading} (MSTR) method proposed as an efficient technique to read long articles, we also try out Longformer~\cite{Longformer}, which is proposed to read long articles efficiently with an attention mechanism that scales linearly with sequence length.  
This attention mechanism is a drop-in replacement for the standard self-attention and combines a local windowed attention with a task motivated global attention.
In this experiment, we replace RoBERTa with Longformer as the text feature extractor. Results are shown in Tab.~\ref{tab:ablation_results}.
Unexpectedly, the Longformer variant of \name underperforms the RoBERTa variant.
The possible reason is that in order to improve training efficiency, Longformer applies local windowed attentions with sparse global attentions.
However, in this task, global attention is needed in every token.
One possible solution for Longformer is to re-do the pretraining with fully global attention. However, this might be a non-trivial task and we will explore this in the future work.

We also conduct experiments to get the best possible number of tokens for MSTR.
We applied number of tokens equal to $512$, $800$, $1000$, $1200$ and $1400$ respectively.
We found that the best choice is $1000$ as it provides nearly the best performance while the training time per epoch is still good enough.

\end{document}